\documentclass[10pt,twocolumn,letterpaper]{article}

\usepackage[pagenumbers]{wacv} 

\usepackage{graphicx}
\usepackage{amsmath}
\usepackage{amssymb}
\usepackage{booktabs}

\usepackage[pagebackref,breaklinks,colorlinks]{hyperref}

\usepackage[capitalize]{cleveref}
\crefname{section}{Sec.}{Secs.}
\Crefname{section}{Section}{Sections}
\Crefname{table}{Table}{Tables}
\crefname{table}{Tab.}{Tabs.}

\def\wacvPaperID{1903} 
\def\confName{WACV}
\def\confYear{2024}

\begin{document}

\title{Hydra: Multi-head Low-rank Adaptation for Parameter Efficient Fine-tuning}

\newcommand\CoAuthorMark{\footnotemark[\arabic{footnote}]}
\newcommand\CorrespondingAuthorMark{\footnotemark[\arabic{footnote}]}
\author{
    Sanghyeon Kim$^1$\thanks{Equal contributions} \and
    Hyunmo Yang$^2$\protect\CoAuthorMark \and
    Younghyun Kim$^2$\protect\CoAuthorMark \and
    Youngjoon Hong$^3$\thanks{Corresponding authors} \and
    Eunbyung Park$^{1,2}\protect\CorrespondingAuthorMark$ \\
    $^1$Department of Electrical and Computer Engineering, Sungkyunkwan University \\
    $^2$Department of Artificial Intelligence, Sungkyunkwan University \\ 
    $^3$Department of Mathematical Sciences, KAIST \\
}
\maketitle

\begin{abstract}
The recent surge in large-scale foundation models has spurred the development of efficient methods for adapting these models to various downstream tasks. Low-rank adaptation methods, such as LoRA, have gained significant attention due to their outstanding parameter efficiency and no additional inference latency. This paper investigates a more general form of adapter module based on the analysis that parallel and sequential adaptation branches learn novel and general features during fine-tuning, respectively. The proposed method, named Hydra, due to its multi-head computational branches, combines parallel and sequential branch to integrate capabilities, which is more expressive than existing single branch methods and enables the exploration of a broader range of optimal points in the fine-tuning process. In addition, the proposed adaptation method explicitly leverages the pre-trained weights by performing a linear combination of the pre-trained features. It allows the learned features to have better generalization performance across diverse downstream tasks. Furthermore, we perform a comprehensive analysis of the characteristics of each adaptation branch with empirical evidence. Through an extensive range of experiments, encompassing comparisons and ablation studies, we substantiate the efficiency and demonstrate the superior performance of Hydra. This comprehensive evaluation underscores the potential impact and effectiveness of Hydra in a variety of applications. Our code is available on \href{https://github.com/extremebird/Hydra}{https://github.com/extremebird/Hydra}
\end{abstract}

\section{Introduction}
\label{sec:intro}

\begin{figure}[ht]
    \centering
    \includegraphics[width=0.4\textwidth]{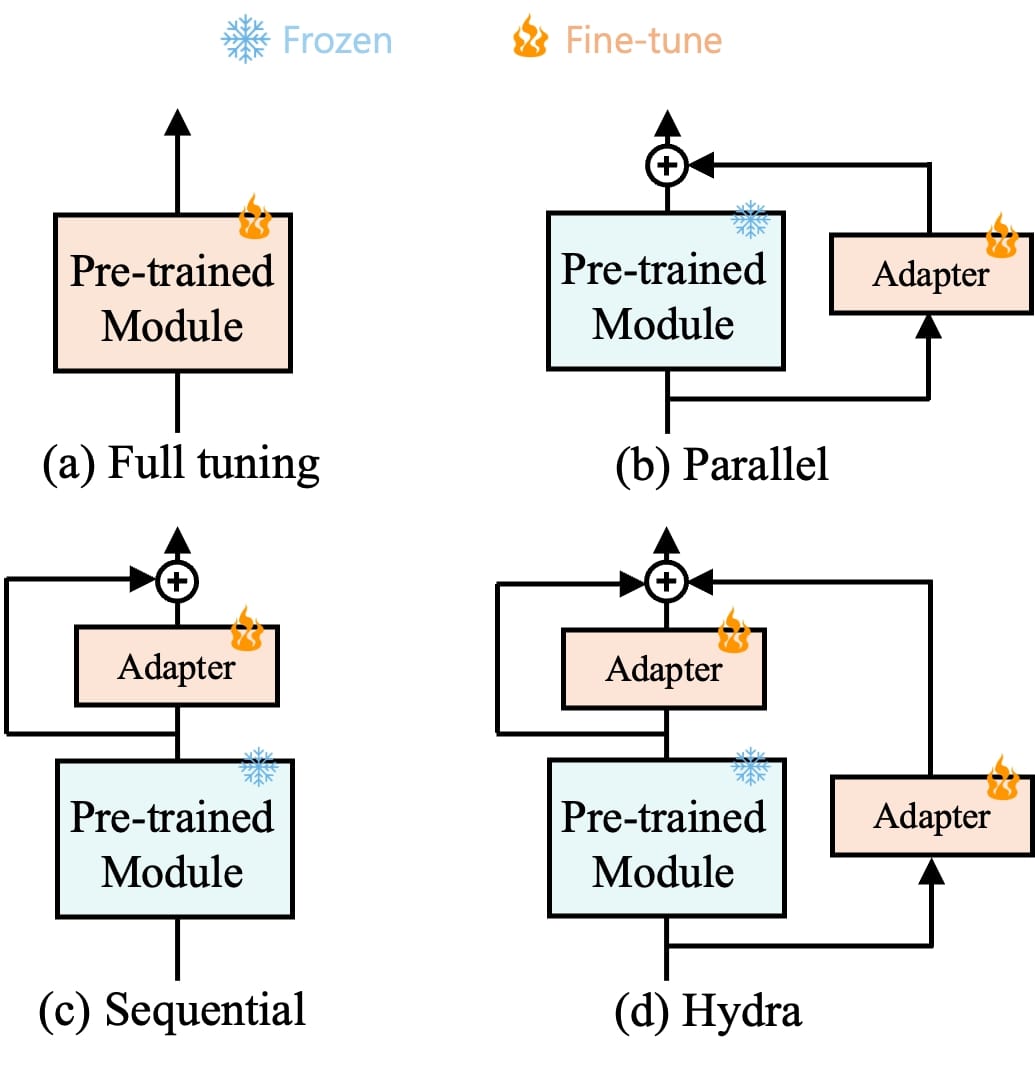}
    \caption{Ghraphical illustration of the model adaptation approaches. (a) fully updates \textit{all} model parameters. (b) and (c) utilize adapter module either in a parallel or a sequential manner. (d) \textit{Hydra} leverages both parallel and sequential adapter.}
    \label{figure:Hydra-figure}
\end{figure}

The large-scale foundation models have been remarkably successful across a broad range of domains and tasks~\cite{brown2020language, kirillov2023segment, raffel2020exploring, devlin2018bert, dehghani2023scaling}. 
Training these large-scale models from scratch is a formidable task, primarily limited to a selected few organizations. The main obstacles hindering broader accessibility are the substantial model sizes, exorbitant computational requirements, and the unavailability of extensive datasets.
In particular, a large model size imposes a significant computational burden even during fine-tuning for downstream tasks.
Efficiently adapting these large-scale models to downstream tasks has emerged as the prevailing practice in numerous applications.

Parameter Efficient Fine-tuning (PEFT) methods~\cite{houlsby2019parameter, hu2021lora, li2021prefix, jia2022visual, chen2022adaptformer, zaken2021bitfit, lian2022scaling} efficiently fine-tune a pre-trained network. Although these methods optimize a significantly smaller number of parameters than the total parameters, they have outperformed the full tuning in various downstream tasks.
Among PEFT methods, adapter-based methods~\cite{houlsby2019parameter, hu2021lora, chen2022adaptformer, lian2022scaling} have demonstrated superior performance and been widely used. They attach lightweight modules, called adapter, to a pre-trained model and optimize only the adapter modules during fine-tuning. Recently, motivated by empirical evidence of low intrinsic dimension in model adaptation, LoRA~\cite{hu2021lora} leverages linear adapter modules to eliminate the additional inference latency existed in previous adapter-based methods~\cite{houlsby2019parameter, chen2022adaptformer}. Furthermore, various matrix factorization techniques have been applied to adapter modules to enhance efficiency~\cite{he2022parameter, zhang2023adaptive}.

While adapter-based methods have become more efficient and advanced, they have been limited to either a parallel or a sequential approach. The parallel (\cref{figure:Hydra-figure}-(b)) and sequential (\cref{figure:Hydra-figure}-(c)) approaches are represented as $f(x) + g(x)$ and $f(x) + g(f(x))$, respectively, where $f$ is a pre-trained module and $g$ is an adapter module. While these two are expressed in a similar fashion, the adapter module of each approach is optimized with different features, $x$ and $f(x)$, as inputs. In other words, task-specific features can be acquired based on how the adapter module is attached during fine-tuning. However, existing adapter-based methods have not extensively explored this aspect.

This paper investigates the characteristics of each attachment approach.
The parallel branch is optimized over the same input $x$ as the pre-trained layer, leading to the learning of task-specific features that have not been pre-trained. This aligns well with the empirical observations, as previous study~\cite{hu2021lora} found that low-rank adaptation often amplifies the important features relevant to specific downstream tasks. On the other hand, the sequential branch learns to combine general features from the pre-trained large-scale model due to its explicit formulation $g(f(x))$.

Based on these characteristics, we propose a more general form of adapter module, called \textit{Hydra}\footnote{Named after the multi-headed Greek mythological beast, rhyme with the original LoRA}, combining parallel and sequential adapter modules. The proposed form is $f(x) + g_p(x) + g_s(f(x))$, where $g_p$ and $g_s$ are parallel and sequential adapter, respectively. This formulation is inherently more expressive than single branch approaches, as it can be reduced to one of them when $g_p(\cdot)=0$ or $g_s(\cdot)=0$.
Our hypothesis is that the introduction of a more general and expressive form enables the exploration of a broader range of local optima. Consequently, this increased flexibility may lead to superior generalization performance for the new tasks.

In addition, we use the linear adapter module of LoRA to construct the proposed method while preserving its advantageous properties. Therefore, the two additional computational branches of our method can be merged after training, hence no additional latency during inference. Moreover, thanks to the simple and versatile linear adapter structure, the proposed module can not only be easily implemented but also be plugged into any linear layers for parameter efficient fine-tuning purposes.

In this paper, we deeply delve into the role of both parallel and sequential branches. 
We observe that each branch learns distinct features during fine-tuning. To elaborate, the parallel branch tends to learn new features by exploring features that were absent during the pre-training phase, and the sequential branch relatively general features by exploiting pre-trained features. The proposed method, \textit{Hydra}, has undergone extensive testing on popular transformer architectures, and we have conducted fine-tuning experiments on diverse datasets spanning vision and natural language tasks. As a result, leveraging both parallel and sequential branches further increases the fine-tuning ability of the model, surpassing other prevalent fine-tuning methods.

\section{Related works}
\subsection{Transformer}
Transformer is a neural network architecture that uses multi-head self attention layers and was initially proposed for machine translation~\cite{vaswani2017attention}. Many large-scale pre-trained transformers~\cite{devlin2018bert, raffel2020exploring, liu2019roberta, brown2020language, radford2019language, lewis2019bart} have exhibited outstanding performance on numerous natural language processing (NLP) tasks, indicating their scalability. These successes of transformers in NLP fields inspired~\cite{dosovitskiy2020image} to introduce the Vision Transformer (ViT), a purely transformer-based backbone architecture for computer vision (CV) tasks that demonstrated promising results. Subsequently, many transformer-based vision models~\cite{touvron2021training, liu2021swin, dong2022cswin, dehghani2023scaling} have been suggested and shown significant improvement on vision tasks, including image classification~\cite{krizhevsky2009learning, deng2009imagenet}, dense prediction~\cite{lin2014microsoft, zhou2017scene}, and image generation~\cite{karras2019style, ho2020denoising}. Moreover, multi-modal training~\cite{radford2021learning} and self-supervised learning~\cite{he2022masked} also accelerate the broad use of ViT. In this paper, we apply our method to transformer architectures widely used in both language and vision tasks.

\subsection{Adapter-based methods}
Adapter-based method is one of the parameter-efficient adaptation methods that involve training only lightweight adapter modules without updating the original parameters of a pre-trained model. \cite{rebuffi2017learning} is a pioneer work that applied adapter modules for multiple vision domain adaptation. Adapter~\cite{houlsby2019parameter} introduced a low-rank residual adapter module that consists of down and up projection with intermediate non-linear function.

Subsequent studies~\cite{ruckle2020adapterdrop, pfeiffer2020adapterfusion, bapna2019simple} demonstrated promising and efficient fine-tuning performance in various NLP tasks. Moreover, Compacter~\cite{karimi2021compacter} leveraged the Kronecker product decomposition and parameter sharing to the projection matrices of adapter modules for more efficient adaptation. VL-adapter~\cite{sung2022vl} successfully applied various adapters to multi-modal (vision and language) tasks, demonstrating their versatility and effectiveness. Adaptformer~\cite{chen2022adaptformer} introduced a parallel adapter to feed-forward networks of ViT for visual recognition tasks. While these adapter tuning methods demonstrated promising results, the additional adapter branches, which have intermediate non-linear functions, slow down the inference speed. 

LoRA~\cite{hu2021lora} proposed a low-rank adaptation module composed solely of linear layers. This design allows the parameters of introduced branches to be merged in an additive manner with the pre-trained parameters at the inference stage, ensuring no latency. Compared to existing adapter tuning, it has shown competitive or even better adaptation ability in NLP fields. AdaLoRA~\cite{zhang2023adaptive} further improved LoRA with singular value decomposition (SVD) for adaptive budget allocation. KAdaptation~\cite{he2022parameter} leveraged a low-rank weight update manner similar to LoRA, in which the update weights are obtained by Kronecker product of shared matrix and low-rank matrix, for fine-tuning vision models. Also, FacT~\cite{jie2023fact} proposed tensorization-decomposition framework, which tensorize whole ViT into a single 3D tensor, then apply Factor-Tuning with various tensor decomposition methods, such as Tensor-Train(TT) or Tucker(TK). SSF~\cite{lian2022scaling} suggested the introduction of scaling and shifting factors to perform linear transformation on features after pre-trained modules of ViT, in order to match the target distribution. Recently, RepAdapter~\cite{luo2023towards} proposed a sequential structural reparameterization scheme for low-rank adaptation modules. These studies have exhibited competitive performance and efficiency, not sagging the inference speed. Building upop these single branch approaches, we leverage parallel and sequential branches together to demonstrate superior performance.

\subsection{Other PEFT approaches}
Besides the success of adapter-based methods, the approaches without adapter have been explored. BitFit \cite{zaken2021bitfit} only trained bias-terms during fine-tuning. Diff-pruning \cite{guo2020parameter} introduced a task-specific diff vector, which is adaptively pruned during training. Token-based tuning approaches~\cite{jia2022visual, lester2021power, li2021prefix, sandler2022fine} also widely used PEFT methods. They involves attaching supplementary tokens, also known as prompt, to the input or intermediate sequence and fine-tune them to guide the model’s attention towards the relevant information for the new task. VPT~\cite{jia2022visual} demonstrated promising performance in vision domain by applying the prompt tuning approach that succeeded in natural language tasks.
While token-based tuning has demonstrated promising tuning ability, the addition of new tokens causes a few drawbacks. It reduces the available input sequence length, potentially limiting the amount of context that the model can process effectively. Moreover, it increases computational complexity. Moreover, applying these approaches to models that utilize local self-attention can pose additional challenges.

\section{Methods}
\subsection{Preliminary}
LoRA~\cite{hu2021lora} applies a linear adapter module on linear (dense) layers of a pre-trained model for efficient model adaptation. It assumes that the intrinsic rank of adaptation matrix $A \in \mathbb{R}^{d \times k}$ is low, allowing a low-rank decomposition on $A$ ($\texttt{rank}(A) \ll \texttt{min}(d, k)$). That is, the adaptation matrix $A$ is decomposed as $A = A_\textrm{up}A_\textrm{down}$, where $A_\textrm{up} \in \mathbb{R}^{d \times r}$ and $A_\textrm{down} \in \mathbb{R}^{r \times k}$ are an up-projection and down-projection adaptation matrix, respectively. Thus, the linear adapter module is formulated as $g(x;A_{up}, A_{down}) = A_\textrm{up}A_\textrm{down}x$. For brevity, henceforth, we use $g(x; A)$ to denote $g(x;A_{up}, A_{down})$


For a given input feature $x \in \mathbb{R}^k$, the forward pass of LoRA is implemented as below:
\begin{align}
    h &= f(x;W_0, b_0) + g(x; A) \label{eq:eq1}\\
    &= W_0x + b_0 + A_\textrm{up}A_\textrm{down}x \label{eq:eq2},
\end{align}
where $h \in \mathbb{R}^d$ is an output vector, $W_0 \in \mathbb{R}^{d \times k}$ is a pre-trained weight matrix, and $b_0 \in \mathbb{R}^d$ is a bias. To efficiently optimize the linear layer during fine-tuning, only the adaptation matrices $A_\textrm{up}$ and $A_\textrm{down}$ are trained, and the pre-trained matrix $W_0$ and bias $b_0$ are frozen.  While using the extra parallel branch $g(x;A)$ is efficient for fine-tuning, it leads latency in inference. Thanks to the linearity, the forward pass in ~\cref{eq:eq2} can be re-implemented as follow:
\begin{align}
    h &= (W_0 + A)x + b_0 \label{eq:eq3}\\
    &= f(x;W_0+A, b_0) \label{eq:eq4},
\end{align}
In other words, during inference, the adapter module can be merged into the pre-trained linear layer, ensuring no additional computational cost.

In ~\cref{eq:eq1}, it is evident that LoRA is one of the parallel approaches. The linear adapter module of LoRA can be optimized without direct dependence on the pre-trained matrix $W_0$. 
Therefore, it would facilitates the ease of learning new features that diverge from the pre-trained features. However, there is a possibility of losing the generalization ability of pre-trained weight matrix.

\subsection{SeqLoRA}
In order to compare parallel and sequential approaches, we introduce SeqLoRA, a sequential form of LoRA, leveraging the idea of a low-rank adaptation on an output vector of the pre-trained linear layer. This results in the following forward pass:
\begin{align}
    h &= f(x;W_0, b_0) + g(f(x;W_0, b_0); B) \label{eq:eq5}\\
    &= W_0x + b_0 + B_\textrm{up}B_\textrm{down}W_0x + B_\textrm{up}B_\textrm{down}b_0 \label{eq:eq6},
\end{align}
where $B \in \mathbb{R}^{d \times d}$ is an adaptation matrix, $B_\textrm{up} \in \mathbb{R}^{d \times r}$ is an up-projection adaptation matrix, and $B_\textrm{down} \in \mathbb{R}^{r \times d}$ is a down-projection adaptation matrix. Similar to LoRA, only the adapter module is optimized and the forward pass for inference can be represented as a single linear layer:
\begin{align}
    h &= (W_0 + BW_0)x + b_0 + Bb_0 \label{eq:eq7}\\
    &= f(x;W_0+BW_0, b_0 + Bb_0) \label{eq:eq8},
\end{align}

We posit that LoRA and SeqLoRA are complementary to each other. SeqLoRA can learn new features for the downstream tasks by linearly combining features from the pre-trained layer. While SeqLoRA has the capacity to learn highly useful features based on the capabilities of the large-scale pre-trained models, it may encounter limitations in learning novel concepts or features that were absent during the pre-training phase.

SeqLoRA shares similarities with the recently proposed RepAdapter~\cite{luo2023towards} in terms of its sequential linear adapter module. However, we introduce it to compare with its parallel counterpart and utilize a component of following our proposed method, \textit{Hydra}.


\subsection{Hydra}

To harness the strengths of both LoRA and SeqLoRA, we introduce \textit{Hydra}, a more general form of linear adaptation module that integrates the capabilities of both methods. \textit{Hydra} allows for the combination and utilization of the advantageous aspects of LoRA and SeqLoRA, providing a comprehensive and flexible framework for efficient and effective model adaptation.
More precisely, it can not only capture novel features easily but also get a broader sight based on general pre-trained features.
For \textit{Hydra}, we combine parallel and sequential adaptation branches, which allows us the following forward pass:
\begin{align}
    h &= f(x;W_0, b_0) + g(x; A ) + g(f(x;W_0, b_0); B) \label{eq:eq9} \\
    &= W_0x + b_0 + A_\textrm{up}A_\textrm{down}x \nonumber \\
    &+ B_\textrm{up}B_\textrm{down}W_0x + B_\textrm{up}B_\textrm{down}b_0 \label{eq:eq10},
\end{align}
where $A \in \mathbb{R}^{d \times k}$, $A_\textrm{up} \in \mathbb{R}^{d \times r_a}$, $A_\textrm{down} \in \mathbb{R}^{r_a \times k}$ is an adaptation matrix for parallel branch and its low-rank decomposition with rank $r_a$, $B \in \mathbb{R}^{d \times d}$, $B_\textrm{up} \in \mathbb{R}^{d \times r_b}$, $B_\textrm{down} \in \mathbb{R}^{r_b \times d}$ is an adaptation matrix for sequential branch and its low-rank decomposition with rank $r_b$. For simplicity, we set rank as $r_a = r_b$ throughout the paper. 
Following LoRA, we use a random Gaussian initialization for down-projection matrices, $A_\textrm{down}$ and $B_\textrm{down}$, and zero initialization for up-projection matrices, $A_\textrm{up}$ and $B_\textrm{up}$. Consequently, at the start of the training, both $A$ and $B$ are initialized to zero. For model adaptation, $A_\textrm{up}$, $A_\textrm{down}$, $B_\textrm{up}$, and $B_\textrm{down}$ are trained based on the gradient descent, while $W_0$ and $b_0$ are not updated.

As depicted in \cref{figure:Hydra-figure}-(d), the implementation for training consists of three branches: pre-trained, parallel, and sequential. After the training, parallel and sequential branches can be merged into the pre-trained branch as follows:
\begin{align}
    h &= (W_0 + A + BW_0)x + b_0 + Bb_0 \label{eq:eq11}\\
    &= f(x;W_0 + A + BW_0, b_0 + Bb_0) \label{eq:eq12},
\end{align}
Therefore, our method does not increase computational complexity during inference.

In addition, it is apparent that LoRA and SeqLoRA can be identified as specific instances of \textit{Hydra} when $B=0$ and $A=0$, respectively. This observation establishes that our method encompasses a more generalized framework for task-specific adaptations. As a result, our approach provides an enhanced modeling capacity to comprehensively capture various adaptation scenarios during fine-tuning.

\subsection{Architecture design}
\label{sec:architecture}

\begin{figure}[t]
  \centering
  \includegraphics[width=0.45\textwidth]{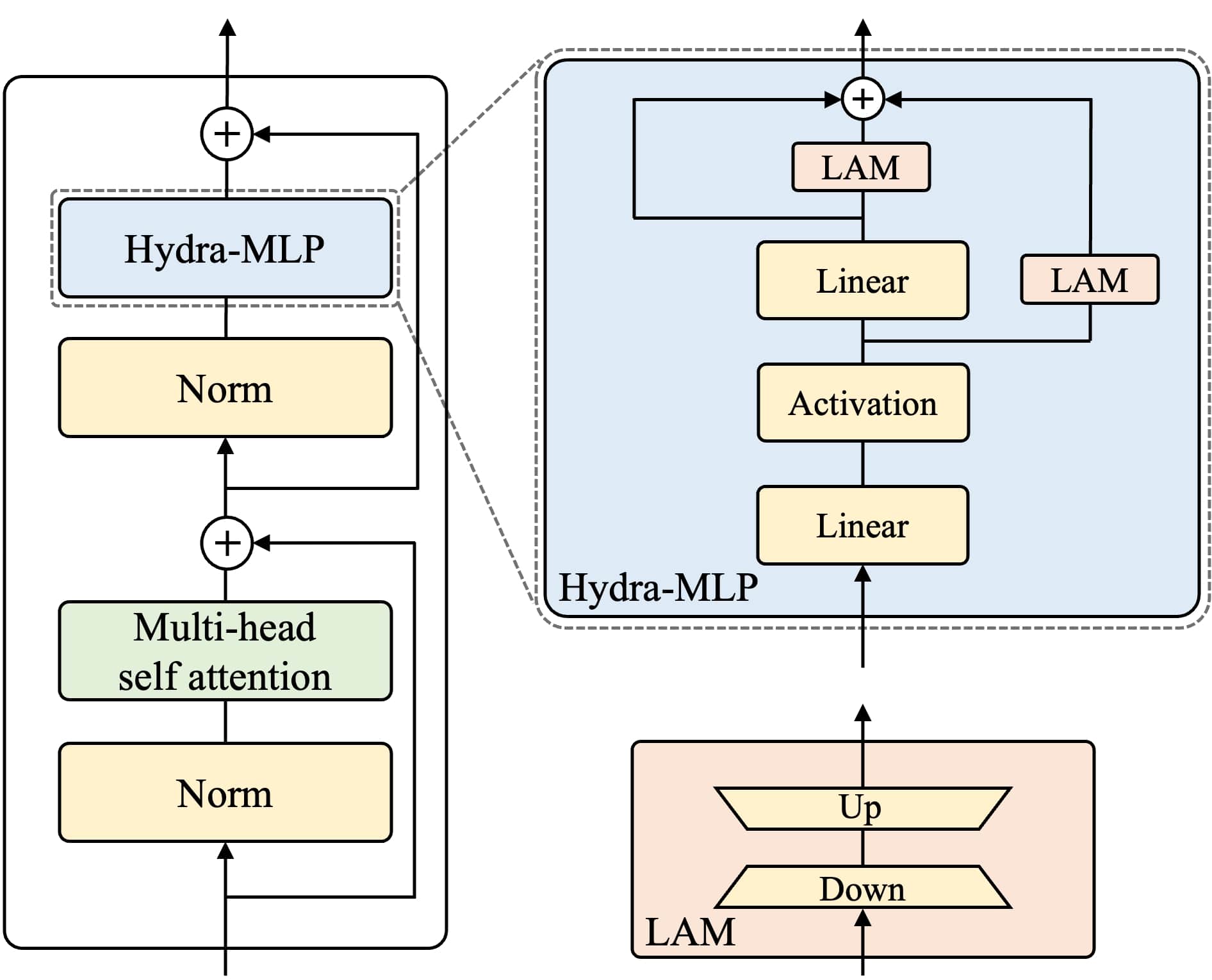}
  \caption{\textit{Hydra}-MLP in a transformer architecture in the training phase. Linear Adapter Module (LAM) implements down projection and up projection on its input in order.}
  \label{figure:Hydra-architecture}
\end{figure}

\begin{table*}[ht]
    \centering
    \setlength{\tabcolsep}{2pt}
    \renewcommand{\arraystretch}{1.3}
    \resizebox{\textwidth}{!}{
    \begin{tabular}{cccc|cccccccccccccccccccc}
        \toprule
        Method & \rotatebox{90}{\#Params (M)} & \rotatebox{90}{Avg Acc. ($\uparrow$)} & \rotatebox{90}{PE ($\uparrow$)} & \rotatebox{90}{Caltech101} & \rotatebox{90}{CIFAR10} & \rotatebox{90}{CIFAR100} & \rotatebox{90}{Country211} & \rotatebox{90}{DTD} & \rotatebox{90}{EuroSAT} & \rotatebox{90}{FER2013} & \rotatebox{90}{FGVCAircraft} & \rotatebox{90}{Food101} & \rotatebox{90}{GTSRB} & \rotatebox{90}{HatefulMemes} & \rotatebox{90}{KITTI-Dist} & \rotatebox{90}{MNIST} & \rotatebox{90}{Flowers102} & \rotatebox{90}{OxfordPets} & \rotatebox{90}{PatchCamelyon} & \rotatebox{90}{SST2} & \rotatebox{90}{Resisc45} & \rotatebox{90}{StanfordCars} & \rotatebox{90}{VOC2007}\\ 
        \midrule
        
        Full tuning & 87.9 & 65.49 & 0.498 &87.64 & \textbf{91.11} & 71.52 & 15.75 & 54.36 & 85.24 & \textbf{52.72} & 26.22 & 83.28 & 74.05 & 55.64 & 39.15 & 65.55 & 80.55 & 87.31 & 64.92 & 59.09 & 75.61 & 57.21 & 82.95\\

        Linear-probing & 0.03 & 66.32 & 0.663 & 90.96 & 90.35 & 67.31 & 17.36 & 62.04 & 72.95 & 51.91 & 29.52 & 83.82 & 56.47 & 55.83 & 40.37 & 77.50 & 92.29 & 88.03 & 59.00 & 59.36 & 78.10 & 68.30 & \textbf{84.99}\\
        \midrule

        Adapter~\cite{houlsby2019parameter} & 1.24 & 65.08 & 0.647 & 90.18 & 90.14 & 73.57 & 16.83 & 57.13 & 67.97 & 41.76 & 30.52 & 83.58 & 58.50	& 48.91	& 37.18 & 80.34	& 90.78	& 86.52	& 59.92	& 58.70 & 79.22 & 67.68 & 82.22\\

        LoRA~\cite{hu2021lora} & 0.18 & 61.48 & 0.614 & 87.64 & 90.52 & 69.69 & 17.12 & 50.16 & 74.03 & 51.04 & 20.01 & 83.76 & 42.96 & 55.88 & 48.05 & 61.36 & 74.28 & 85.49 & 63.20 & 57.04 & 62.09 & 54.89 & 80.33\\

        Compacter~\cite{karimi2021compacter} & 0.08 & 62.79 & 0.628 & 89.02 & 79.96 & 44.33 & \textbf{28.22} & 52.93 & 50.48 & 35.46 & \textbf{41.13} & 78.28 & 66.90 & 47.60	& \textbf{57.72} & 85.82 & 88.29 & 79.23 & 61.83 & \textbf{64.22} & 63.76 & 64.79 & 75.84\\

        Kadaptation~\cite{he2022parameter} & 0.08 & 68.92 & 0.689 & 88.96	& 90.03	& 73.92 & 17.53 & 63.97 & 76.25 & 47.45 & 30.04 & \textbf{84.38}	& 80.71	& 55.86	& 42.29	& 85.20 & \textbf{93.19} & 89.05 & 63.39 & 59.18 & 79.96 & 70.21 & 84.49\\

        \midrule
        Hydra & 0.20 & \textbf{70.95} & \textbf{0.709} & \textbf{91.23} &90.89 	&\textbf{74.20} &17.75 	&\textbf{64.47} 
        &\textbf{87.00} &51.10 	&33.05 	&84.27 	&\textbf{87.11} &\textbf{55.91} &42.05 	&\textbf{90.76} &93.18 	&\textbf{89.38} &\textbf{70.83}	&59.58 	&\textbf{82.41} &\textbf{71.19} &82.66 \\
        \bottomrule
        
    \end{tabular}}
    \caption{ELEVATER experiment results. We use CLIP pre-trained ViT-Base-224/32 as a backbone model.} 
    \label{table:clip-few}
\end{table*}

While our approach is designed to be compatible with any linear layers, in this work, we focus on its application to the MLP blocks within transformer architectures, which have been widely used in recent large-scale models. As illustrated in ~\cref{figure:Hydra-architecture}, a typical transformer block consists of a multi-head self attention (MSA) block and an MLP block, interleaved by non-linear activations and layer normalization. We replace the proposed adaptation module with the last layer of the MLP block. Since non-linear activations are not employed in the final linear layer of the MLP, we can avoid the potential `additional latency' that could arise during the inference. We refer to the MLP block, to which our method is applied, as \textit{Hydra}-MLP.

In addition, this design choice is also motivated by recent studies, revealing that self-attention blocks in the transformers tend to diminish high-frequency information, whereas MLP blocks amplify it~\cite{park2022vision, wang2022anti}. As \textit{Hydra}-MLP contains SeqLoRA that is designed to exploit the pre-trained features through the linear combination, our approach effectively encourages the model to promote useful high-frequency features for specific downstream tasks. Unless specified, in this paper, \textit{Hydra} indicates \textit{Hydra}-MLP.

\section{Experimental results}

We substantiate the versatility of our method \textit{Hydra} through an extensive range of experiments including both vision and natural language tasks. Next, we analyze the characteristics based on the utilization approach of adapter branch and discuss the computational efficiency of the proposed method. Finally, we verify the effectiveness of our architecture design by conducting ablation studies.
\subsection{Few-shot experiments}
\label{subsection:fewshot}
First, since various fine-tuning applications opt to fall down to the condition of limited data accessibility, we validated our proposed \textit{Hydra} in a few-shot learning scenario using 20 image classification datasets from the ELEVATER benchmark~\cite{li2022elevater}. Each dataset comprises a distinct number of labels along their corresponding images. Following the previous work~\cite{he2022parameter}, we used the CLIP pre-trained ViT-Base-224/32 as a backbone model. And, we set the bottleneck rank of \textit{Hydra} as $r_a = r_b = 2$. Detailed experimental settings and statistics of each dataset are reported in \cref{appxB,appxC}, respectively.

As shown in ~\cref{table:clip-few}, \textit{Hydra} achieved the highest accuracy score on 11 out of 20 datasets and surpassed the other PEFT methods with regard to the average accuracy. Furthermore, we report the PE score~\cite{li2022elevater} to compare accuracy-efficiency trade-off. The PE score is defined as follow:
\begin{align}
    \text{PE} = \textrm{accuracy}\cdot \exp(-\log_{10}(p/M_0 + 1)),
\end{align}
where $p$ is the number of trainable parameters, and $M_0$ is a magnitude of pre-trained model's parameters. We set $M_0 = 10^8$. We observed that our method also attained the highest PE score in ~\cref{table:clip-few}. As a result, the proposed method is not only effective but also efficient for few-shot learning. 
\subsection{VTAB-1k experiments}
\label{subsection:vtab}
\begin{table*}[ht!]
  \centering
  \setlength{\tabcolsep}{2pt}
  \renewcommand{\arraystretch}{1.3}
  {\small
  \begin{tabular}{ccc|ccccccc|cccc|cccccccc}
    \toprule
        \multicolumn{3}{c}{} & \multicolumn{7}{|c|}{\textbf{Natural}} & \multicolumn{4}{|c|}{\textbf{Specialized}} & \multicolumn{8}{|c}{\textbf{Structured}} \\
        Method & \rotatebox{90}{\#Params (M)} & \rotatebox{90}{Avg Acc.} & \rotatebox{90}{CIFAR100} & \rotatebox{90}{Caltech101} & \rotatebox{90}{DTD} & \rotatebox{90}{Flowers102} & \rotatebox{90}{OxfordPets} & \rotatebox{90}{SVHN} & \rotatebox{90}{Sun397} & \rotatebox{90}{PatchCamelyon} & \rotatebox{90}{EuroSAT} & \rotatebox{90}{Resisc45} & \rotatebox{90}{Retinopathy} & \rotatebox{90}{Clevr-Count} & \rotatebox{90}{Clevr-Dist} & \rotatebox{90}{DMLab} & \rotatebox{90}{KITTI-Dist} & \rotatebox{90}{dSPR-Loc} & \rotatebox{90}{dSPR-Ori} & \rotatebox{90}{sNORB-Azim} & \rotatebox{90}{sNORB-Ele} \\
    \midrule
        Full tuning & 85.8 & 68.9 & 68.9  & 87.7  & 64.3 & 97.2 & 86.9 & 87.4 & 38.8 & 79.7 & 95.7 & 84.2 & 73.9 &56.3 & 58.6 & 41.7 & 65.5 & 57.5 & 46.7 & 25.7 & 29.1\\
        Linear-probing & 0.04 & 57.6 & 64.4 & 85.0 & 63.2 & 97.0 & 86.3 & 36.6 & 51.0 & 78.5 & 87.5 & 68.5 & 74.0 & 34.3 & 30.6 & 33.2 & 55.4 & 12.5 & 20.0 & 9.6 & 19.2\\
    \midrule
        Adapter~\cite{houlsby2019parameter} & 0.16 & 73.9 & 69.2 & 90.1 & 68.0 & 98.8 & 89.9 & 82.8 & 54.3 & 84.0 & 94.9 & 81.9 & 75.5 & 80.9 & 65.3 & 48.6 & 78.3 & 74.8 & 48.5 & 29.9 & 41.6\\
        
        AdaptFormer~\cite{chen2022adaptformer} & 0.16 & 74.7 & 70.8 & 91.2 & 70.5 & 99.1 & 90.9 & 86.6 & 54.8 & 83.0 & 95.8 & 84.4 & \textbf{76.3} & 81.9 & 64.3 & 49.3 & 80.3 & 76.3 & 45.7  & 31.7 & 41.1 \\
        
        LoRA~\cite{hu2021lora} & 0.29 & 74.5 & 67.1 & 91.4 & 69.4 & 98.8 & 90.4 & 85.3 & 54.0 & 84.9 & 95.3 & 84.4 & 73.6 & 82.9 & \textbf{69.2} & 49.8 & 78.5 & 75.7 & 47.1 & 31.0 & 44.0 \\
        
        VPT~\cite{jia2022visual}  & 0.53 & 72.0 & \textbf{78.8} & 90.8 & 65.8 & 98.0 & 88.3 & 78.1 & 49.6 & 81.8 & 96.1 & 83.4 & 68.4 & 68.5 & 60.0 & 46.5 & 72.8 & 73.6 & 47.9 & 32.9 & 37.8\\
        
        NOAH~\cite{zhang2022neural} & 0.36 & 75.5 & 69.6 & \textbf{92.7} & 70.2 & 99.1 & 90.4 & 86.1 & 53.7 & 84.4 & 95.4 & 83.9 & 75.8 & 82.8 & 68.9 & 49.9 & 81.7 & \textbf{81.8} & 48.3 & 32.8 & \textbf{44.2}\\
        
        SSF~\cite{lian2022scaling} & 0.22 & 75.7 & 69.0 & 92.6 & \textbf{75.1} & \textbf{99.4} & \textbf{91.8} & 90.2 & 52.9 & \textbf{87.4} & 95.9 & \textbf{87.4} & 75.5 & 75.9 & 62.3 & \textbf{53.3} & 80.6 & 77.3 & \textbf{54.9} & 29.5 & 37.9\\
        
        FacT~\cite{jie2023fact} & 0.07 & 75.6 & 70.6 & 90.6 & 70.8 & 99.1 & 90.7 & 88.6 & 54.1 & 84.8 & \textbf{96.2} & 84.5 & 75.7 & 82.6 & 68.2 & 49.8 & 80.7 & 80.8 & 47.4 & 33.2 & 43.0\\
        
        RepAdapter~\cite{luo2023towards} & 0.22 & 76.1 & 72.4 & 91.6 & 71.0 & 99.2 & 91.4 & \textbf{90.7} & 55.1 & 85.3 & 95.9 & 84.6 & 75.9 & 82.3 & 68.0 & 50.4 & 79.9 & 80.4 & 49.2 & \textbf{38.6} & 41.0\\
    \midrule
        Hydra & 0.28 & \textbf{76.5} & 72.7 & 91.3 & 72.0 & 99.2 & 91.4 & \textbf{90.7} & \textbf{55.5} & 85.8 & 96.0 & 86.1 & 75.9 & \textbf{83.2} & 68.2 & 50.9 & \textbf{82.3} & 80.3 & 50.8 & 34.5 & 43.1\\
    \bottomrule
  \end{tabular}}
  \caption{VTAB-1k experiment results. All methods are conducted by ViT-Base-224/16 pre-trained on ImageNet-21k.}
  \label{table:vtab}
\end{table*}
Next, we conducted the experiment on the VTAB-1k benchmark~\cite{zhai2019visual} to compare \textit{Hydra} with the state-of-the-art PEFT methods. The VTAB-1k benchmark consists 19 vision datasets and each dataset are categorized into three groups with different concepts, i.e., the \textit{Natural}, \textit{Specialized} and \textit{Structured}. 
We used ViT-Base-224/16 model pre-trained on ImageNet-21k in a supervised manner. Following prior works~\cite{luo2023towards}, we applied our \textit{Hydra} module on every layer of both projection layer in attention block and final linear layer of MLP block with low-rank dimension $r_a = r_b = 2$ for this experiment. More details of the experiments are in \cref{appxB,appxC}.

We noted that \textit{Hydra} excels recent PEFT methods in ~\cref{table:vtab}. Compared to existing non-linear adapter methods~\cite{houlsby2019parameter, chen2022adaptformer}, our method has demonstrated enhanced performance, avoiding any additional inference latency through the combination of linear operations. Therefore this signifies linear adapter modules can also function well in a multi-branch approach.

Furthermore, it is noteworthy the proposed method, which combines parallel and sequential adaptation branches, outperforms previous single branch (either parallel or sequential) approaches~\cite{hu2021lora, lian2022scaling, jie2023fact, luo2023towards}. To effectively learn task-specific feature during fine-tuning, in other words, both a parallel branch that learns novel concepts and a sequential branch that transforms pre-trained features need to be used in conjunction. As a result, introducing a more comprehensive and expressive structure is helpful for proficient task adaptation.
\begin{table}[t]
  \centering
  \setlength{\tabcolsep}{2pt}
  \renewcommand{\arraystretch}{1.3}
  \resizebox{0.45\textwidth}{!}{
  \begin{tabular}{ccc|cccccccc}
    \toprule
        Method & \rotatebox{90}{\#Params (M)}& \rotatebox{90}{Avg.} &\rotatebox{90}{MNLI} & \rotatebox{90}{SST-2} & \rotatebox{90}{MRPC} & \rotatebox{90}{CoLA} & \rotatebox{90}{QNLI} & \rotatebox{90}{QQP} & \rotatebox{90}{RTE} & \rotatebox{90}{STS-B}  \\
    \midrule
        Full tuning         & 125 & 86.4 &  \textbf{87.6} & 94.8  & 90.2  &  63.6 & 92.8 & \textbf{91.9} & 78.7 & 91.2 \\
        BitFit~\cite{zaken2021bitfit}              & 0.1   & 85.2 &  84.7 & 93.7  & \textbf{92.7}  &  62.0 & 91.8 & 84.0 & 81.5 & 90.8 \\
        AdapterDrop~\cite{ruckle2020adapterdrop}         & 0.3   & 84.4 &  87.1 & 94.2  & 88.5  &  60.8 & 93.1 & 90.2 & 71.5 & 89.7 \\
        AdapterDrop~\cite{ruckle2020adapterdrop}         & 0.9   & 85.4 &  87.3 & 94.7  & 88.4  &  62.6 & 93.0 & 90.6 & 75.9 & 90.3 \\
    \bottomrule
        LoRA~\cite{hu2021lora}                & 0.3   & 87.2 &  87.5 & \textbf{95.1}  & 89.7  &  63.4 & \textbf{93.3} & 90.8 & 86.6 & 91.5 \\
        Hydra               & 0.3   & \textbf{87.9}&  87.5 & 95.0  & 92.2  &  \textbf{65.4} & 92.8 & 90.8 & \textbf{87.4} & \textbf{91.7} \\
    \bottomrule
  \end{tabular}}
  \caption{Natural language understanding results. We report the Matthew's correlation for CoLA, Pearson correlation for STS-B, and accuracy for the others.}
  \label{table:Table3}
\end{table}

\subsection{Natural language understanding experiments}
\label{subsection:nlu}

In the field of NLP, transformers have achieved great success, leading numerous large-scale pre-trained transformer models. Hence, many PEFT methods were initially proposed for NLP tasks. Therefore, in this section, we validated that our method can effectively fine-tune a pre-trained NLP model as well.
We performed natural language understanding experiments on GLUE benchmark~\cite{wang2019superglue}. Following~\cite{hu2021lora}, we used the pre-trained RoBERTa (base)~\cite{liu2019roberta}, which originally has 125M trainable parameters from the HuggingFace Transformers library~\cite{wolf2020transformers}. More experimental details are in \cref{appxB,appxC}.

As shown in ~\cref{table:Table3}, \textit{Hydra} has demonstrated superior adaptation capability compared to full tuning while requiring significantly fewer trainable parameters. Similarly, akin to results from vision task experiments, the proposed method outperforms existing PEFT approaches. Notably, despite both LoRA and \textit{Hydra} applying the same linear adapter module, \textit{Hydra} achieves a substantial lead over LoRA, showing a significant performance advantage (+0.7 on average). This underscores the potential of the proposed method in NLP tasks as well. In essence, our multi-branch adapter module exhibits strong performance across domains, making it versatile and applicable in various fine-tuning scenarios.

\subsection{Analysis}
\label{subsection:analysis}
Adapter-based methods can be categorized into parallel and sequential approaches based on the attachment manner. While the formulations (~\cref{eq:eq1,eq:eq5}) are similar, they are trained in distinct way because of different input features. The parallel branch learns new features by exploring features that were absent during the pre-training phase. On the other hand, the sequential branch learns relatively general features by exploiting pre-trained features. In this section, we delve into the attributes of the parallel and sequential branches with experimental evidence. Following that, we also analyze \textit{Hydra} from the perspective of efficiency, an important element of the PEFT methods.

\textbf{Subspace similarity of weight matrix}
We analyze each branch in terms of the weight matrix. To do so, we measured the similarity between pre-trained weight matrix $W_0$ and the weight matrices of each branch. In ~\cref{eq:eq11}, the weight matrices of the parallel and sequential branches are represented as $A$ and $BW_0$, respectively, for input $x$. Following ~\cite{hu2021lora}, we leveraged subspace similarity defined as below:
\begin{align}
    \phi(M, N, i, j) = \frac{\lVert {U^i_M}^T U^j_N \rVert^2_F}{min\{i,j\}},
\end{align}
where matrices $U^i_M \in \mathbb{R}^{d \times i}$ and $U^j_N \in \mathbb{R}^{d \times j}$ are formed by extracting from first to i-th and j-th columns of the left singular matrix of matrices $M$ and $N$, respectively. We evaluated the similarity between the top 10$\%$ singular directions in the pre-trained matrix $W_0$ and the top 2 singular directions in the adaptation weight matrix $A$ or $BW_0$.

\begin{figure}
    \centering
    \includegraphics[width=0.45\textwidth]{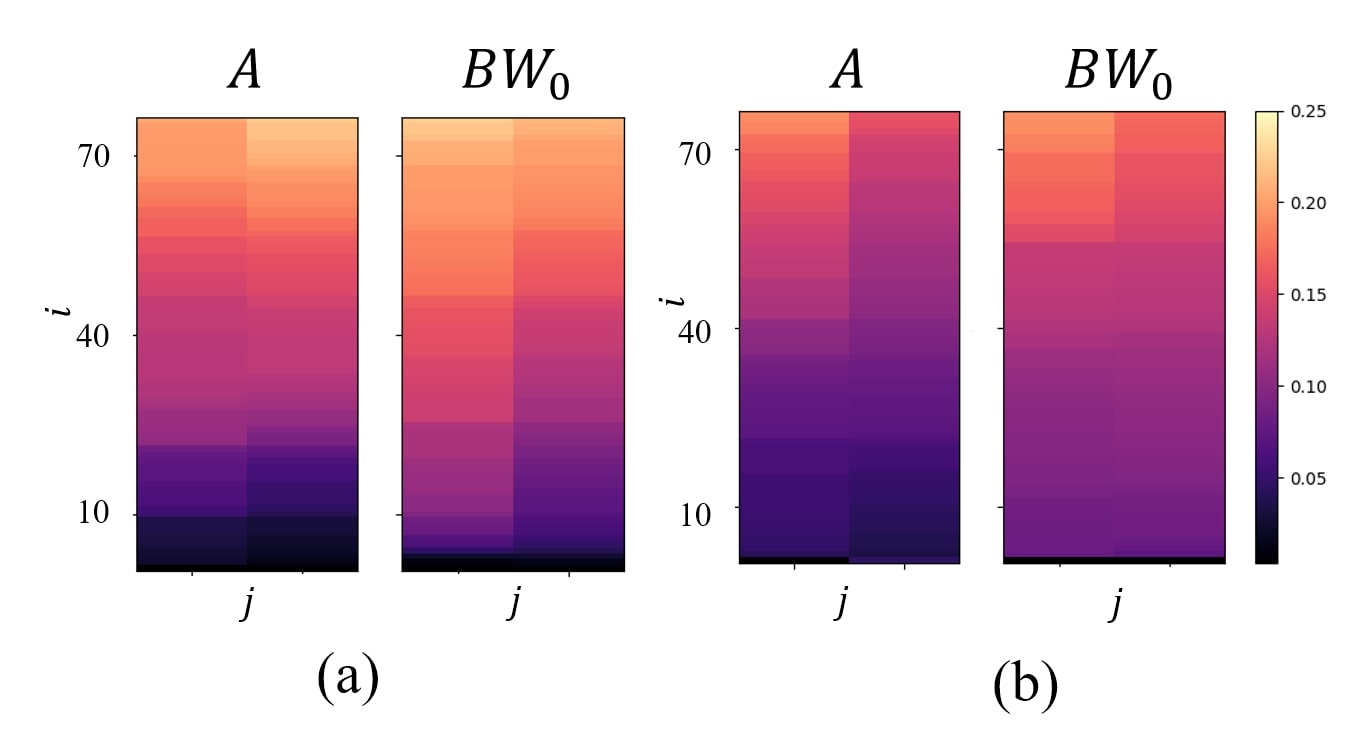}
    \caption{Normalized subspace similarity. $\phi(W_0, A, i, j)$ and $\phi(W_0, BW_0, i, j)$ of each transformer block trained on CIFAR100 in VTAB-1k benchmark. (a) and (b) correspond to the 6th and 9th blocks of ViT-Base-224/16, respectively.}
    \label{figure:similarity}
\end{figure}

In ~\cref{figure:similarity}, we observed higher overall similarity values between $BW_0$ and $W_0$ compared to $A$ and $W_0$ due to explicitly leveraging $W_0$. This indicates the sequential branch tends to learn general features that are relatively similar to pre-trained features. Furthermore, the majority of similarity values do not exceed 0.25 for both $A$ and $BW_0$. It implies the \textit{Hydra} module enhances task-specific features instead of the previously amplified features by $W_0$. Thus, our multi-branch module effectively fulfills the role of the adapter module, which needs to learn task-specific features.

\textbf{Visualization of the feature space}
We conducted t-SNE~\cite{van2008visualizing} visualization on the embedding features of the \texttt{[CLS]} token in the last transformer block after fine-tuning. In this visualization, we visualized the embedding feature, distinguishing it into pre-trained branch output $f(x;W_0, b_0)$, parallel branch output $g(x;A)$, and sequential branch output $g(f(x;W_0, b_0);B)$. Based on this, we interpreted what features each branch is trained to represent.

\begin{figure}
    \centering
    \includegraphics[width=0.45\textwidth]{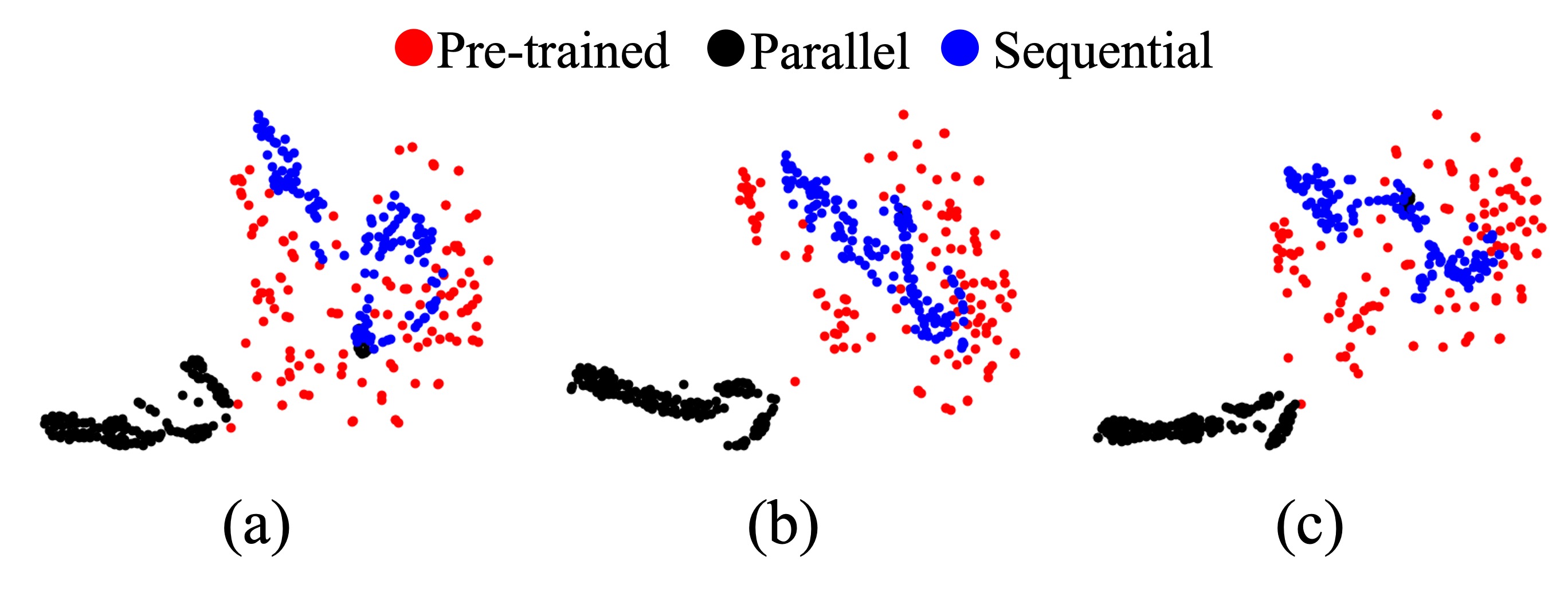}
    \caption{t-SNE visualizations of \texttt{[CLS]} token embedding. We used ViT-Base-224/32 trained on CIFAR100 in VTAB-1k benchmark. (a), (b), and (c) are the results of different random seeds, respectively. For each result, we used 128 images as input.}
    \label{figure:tsne}
\end{figure}

As illustrated in ~\cref{figure:tsne}, we can observe a notable difference in the distribution of output features between the parallel branch and the sequential branch. This clearly demonstrates that each branch hold distinct characteristics. In particular, the distribution of output features in the sequential branch primarily within the feature space of the pre-trained branch. It indicates that the sequential branch learns features that are similar to well-generalized pre-trained ones. On the other hand, the parallel branch learns the unique features that were not acquired during pre-training.

\begin{table}[ht]
  \centering
  \setlength{\tabcolsep}{2pt}
  \begin{tabular}{c|ccc}
    \toprule
        Method & \#Params (M) &  \begin{tabular}[c]{@{}c@{}}Training Time\\ (sec/epoch)\end{tabular}\\
    \midrule
        Full tuning    & 85.8     & 148.95\\
        LoRA           & 0.22     & 115.44\\
        SeqLoRA        & 0.21     & 112.19\\
        Hydra          & 0.21     & 119.37\\
    \bottomrule
  \end{tabular}
  \caption{Training time comparison of full tuning, single branch approaches, and \textit{Hydra} with CIFAR10 on a single RTX3090 GPU. The results are
        tested with a batch size of 128 and input resolution of 224 × 224. Reported training time(sec/epoch) is the average across 10 epochs.}
  \label{table:training_complexity}
\end{table}

\begin{table*}[ht!]
    \begin{minipage}[t]{0.3\textwidth}
        \centering
        \setlength{\tabcolsep}{2pt}
        \renewcommand{\arraystretch}{1.3}
        \begin{tabular}{c|cc}
            \toprule
            Method & \#Params (M) & Avg Acc. \\
            \midrule
            LoRA   & 0.31  & 75.6 \\
            SeqLoRA & 0.28  & 75.4 \\
            Hydra & 0.28  & \textbf{76.5} \\
            \bottomrule
        \end{tabular}
        \caption{Comparison of LoRA, SeqLoRA, and \textit{Hydra} on VTAB-1K benchmark. Each method is applied to both MSA and MLP blocks.} 
        \label{table:ablation_Hydra_architecture}
    \end{minipage}
    \hfill
    \begin{minipage}[t]{0.3\textwidth}
        \centering
        \setlength{\tabcolsep}{2pt}
        \renewcommand{\arraystretch}{1.3}
        \begin{tabular}{c|cc}
            \toprule
                Method  & \#Params (M) & Avg. \\
            \midrule
                LoRA & 0.3   &  87.2\\
                SeqLoRA  & 0.3   &  87.4\\
                Hydra & 0.3 & \textbf{87.9}\\
            \bottomrule
        \end{tabular}
        \caption{Comparison of LoRA, SeqLoRA, and \textit{Hydra} on GLUE benchmark. The reported numbers represent average performance. Each method is applied to MLP blocks.}
        \label{table:Table6}
    \end{minipage}
    \hfill
    \begin{minipage}[t]{0.3\textwidth}
    \centering
    \setlength{\tabcolsep}{2pt}
    \renewcommand{\arraystretch}{1.3}
    \begin{tabular}{c|cc}
        \toprule
        Block & \#Params (M) & Avg Acc. \\
        \midrule
        MSA   & 0.20  & 70.45  \\
        MLP         & 0.20  & \textbf{70.95}  \\
        \bottomrule
    \end{tabular}
    \caption{Ablation results for the optimal position of \textit{Hydra} module on the ELEVATER benchmark.}
    \label{table:ablation_Hydra_position}
    \end{minipage}
\end{table*}

\textbf{Computational efficiency}
Here, we address the parameter-efficiency of our \textit{Hydra} with computational complexity. For simplicity, we assume the input and output of linear adapter module have the same dimension $d$. Then, the linear adapter module has a computational complexity of $O(2rd)$. This is because it is defined as $g(x;A) = A_\textrm{up}A_\textrm{down}x$, where $A_{up} \in \mathbb{R}^{d \times r}$, and $A_{down} \in \mathbb{R}^{r \times d}$. Therefore, the computational complexity of the single branch methods, LoRA and SeqLoRA, is $O(2rd)$. For $r_a = r_b = r$, \textit{Hydra} fundamentally has two branches, leading to an increase in computational complexity. However, in all our experiments, we set $r_a = r_b = \frac{r}{2}$, resulting in both time and memory complexity of $O(2rd)$. It implies that the computational complexity of LoRA, SeqLoRA, and \textit{Hydra} are theoretically identical.


However, when applied to the real applications, multi-branch design of \textit{Hydra} is likely to lead bottlenecks on GPU. To figure out the bottlenecks, we compared the training time of each method on CIFAR10 dataset. For a fair comparison, we applied all methods, including \textit{Hydra}, to MLP blocks. We used ViT-Base-224/32 model with batch size of 128.
The results are shown in ~\cref{table:training_complexity}. 
It demonstrates that when the number of parameters, meaning memory consumption, is similar, it can be observed that single branch methods are generally faster than \textit{Hydra}.
However, the difference is not significant, and as observed in previous experiments, \textit{Hydra} has demonstrated exceptional adaptation performance compared to other methods.
Additionally, \textit{Hydra} has the advantage of no additional inference latency. Therefore, the adaptation branches used for fine-tuning have no impact on the inference computational complexity.
\subsection{Ablation studies}
\label{subsection:ablation}

In this section, we performed ablation studies to validate the rationale behind our architecture design. First, we carried out a head-to-head comparison to fairly assess the efficacy of our approach. Next, we verified the effective position of \textit{Hydra} in the transformer architecture. Here, we present only the summarized tables,
~\cref{table:ablation_Hydra_architecture,table:Table6,table:ablation_Hydra_position}. The full tables are reported in \cref{appxA}.

\textbf{Head-to-head comparison}
\textit{Hydra} is a method that combines parallel and sequential branches, i.e., LoRA and SeqLoRA. We performed experiments, ablating one of these branches, to prove the advantage of combined branch method. To do so, we leveraged the vision experiments in \cref{subsection:vtab} and the natural language experiments in \cref{subsection:nlu} with same experimental settings. For fair comparison, we applied each method to the blocks where the adapter module was attached in each experiment. We configured the low-rank $r$ to ensure a similar number of trainable parameters.

As shown in \cref{table:ablation_Hydra_architecture,table:Table6}, \textit{Hydra} exhibits the highest average performance in both experiments. In addition, while there is no significant difference in performance between LoRA and SeqLoRA, \textit{Hydra} demonstrates a notable discrepancy. This observation implies that combining LoRA and SeqLoRA, \textit{Hydra}, is more effective way than using each. Considering our analysis that parallel and sequential branch are complementary in nature, the proposed method can be regarded as effectively integrating the strengths of each branch.
Consequently, the general and expressive form of our method enables outstanding fine-tuning across diverse task domains, regardless of the specific domain.

\textbf{Position of Hydra module}
Inherently, \textit{Hydra} module can be applied to any linear layer of the transformer, such as projection layers of MSA blocks or linear layers that reside in MLP blocks. We mainly applied the \textit{Hydra} module to MLP blocks, guided by the distinctive attribute of each block. To make it more concretely, we empirically investigate the optimal block for \textit{Hydra} module. It was performed on the ELEVATER benchmark experiments in 
\cref{subsection:fewshot}. 

The results are shown in \cref{table:ablation_Hydra_position}. We observed that when applying \textit{Hydra} module to MLP blocks, it demonstrates better performance. Therefore, our architecture design, \textit{Hydra}-MLP described in \cref{figure:Hydra-architecture}, is reasonable. Furthermore, it indicates that our method well transforms pre-trained features with amplified high-frequency by MLP block into task-specific features.



\section{Conclusion}

In this paper, we conduct an in-depth analysis of the roles of each adaptation branch, parallel and sequential, which has not been explored. We demonstrate that parallel branch inclines towards acquiring novel features through the exploration of absent features during pre-training phase, while the sequential branch utilizes pre-trained features to capture relatively general features.

We also propose a general and expressive adaptation formulation, \textit{Hydra}, which combines parallel and sequential adaptation branch to integrate the capabilities of both branches. By leveraging linear adapter module, it has no additional inference latency and can be applied to any linear layer. Furthermore, thanks to its simple structure, \textit{Hydra} can be easily implemented. The proposed method demonstrates superior performance on comprehensive experiments, including both vision and natural language tasks, without bells and whistles. This shows the versatility of \textit{Hydra} in fine-tuning applications.

Since our focus is primarily on analyzing the characteristics of each branch and demonstrating the effectiveness of the multi-branch approach, we utilize simple linear adapter module. However, the form of \textit{Hydra} is not influenced by the adapter module. Therefore, existing adapter-based methods can be easily extended to multi-branch variants. We anticipate that the general and expressive formation of \textit{Hydra} will be widely adopted in the field of the parameter efficient fine-tuning.

{\small
\bibliographystyle{ieee_fullname}
\bibliography{egbib}
}


\clearpage
\appendix
\renewcommand\thesection{\Alph{section}}
\setcounter{section}{0}

\twocolumn[
\begin{center}
\Large{\bf{Hydra: Multi-head Low-rank Adaptation for Parameter Efficient Fine-tuning\\ Appendix}}\par\vspace{3ex}
\end{center}]

\section{Detailed results of ablation studies}\label{appxA}
Due to constraints of space, we reported only the summarized tables in the main paper. Here, we present full results for each experiment.

\begin{itemize}
\item \cref{table:vtab_appx}: The detailed results of head-to-head comparison on VTAB-1k benchmark.
\item \cref{table:nlu_ablation}: The detailed results of head-to-head comparison on GLUE benchmark.
\item \cref{table:position_hydra}: The detailed results of the optimal position of \textit{Hydra} module experiment on ELEVATER benchmark.
\end{itemize}

\begin{table*}[ht]
  \centering
  \setlength{\tabcolsep}{2pt}
  \renewcommand{\arraystretch}{1.3}
  {\small
  \begin{tabular}{ccc|ccccccc|cccc|cccccccc}
    \toprule
        \multicolumn{3}{c}{} & \multicolumn{7}{|c|}{\textbf{Natural}} & \multicolumn{4}{|c|}{\textbf{Specialized}} & \multicolumn{8}{|c}{\textbf{Structured}} \\
        Method & \rotatebox{90}{\#Params(M)} & \rotatebox{90}{Avg Acc.} & \rotatebox{90}{CIFAR100} & \rotatebox{90}{Caltech101} & \rotatebox{90}{DTD} & \rotatebox{90}{Flowers102} & \rotatebox{90}{OxfordPets} & \rotatebox{90}{SVHN} & \rotatebox{90}{Sun397} & \rotatebox{90}{PatchCamelyon} & \rotatebox{90}{EuroSAT} & \rotatebox{90}{Resisc45} & \rotatebox{90}{Retinopathy} & \rotatebox{90}{Clevr-Count} & \rotatebox{90}{Clevr-Dist} & \rotatebox{90}{DMLab} & \rotatebox{90}{KITTI-Dist} & \rotatebox{90}{dSPR-Loc} & \rotatebox{90}{dSPR-Ori} & \rotatebox{90}{sNORB-Azim} & \rotatebox{90}{sNORB-Ele} \\
    \midrule
        LoRA & 0.31 & 75.6 & \textbf{72.86} & 90.5 & \textbf{73.2} & \textbf{99.3} & \textbf{91.4} & 88.4 & 55.0 & 85.6 & 94.4 & 85.9 & 74.4 & 81.5 & \textbf{68.2} & 49.4 & 78.6 & 78.5 & 49.5 & 32.7 & 43.0\\
        SeqLoRA & 0.28 & 75.4 & 72.4 & 89.2 & 71.4 & 99.0 & 90.8 & 87.7 & 55.2 & 85.6 & 94.3 & 84.9 & 74.7 & 82.7 & 67.7 & 49.3 & 79.9 & 81.0 & 50.1 & 32.8 & 40.3\\
        Hydra & 0.28 & \textbf{76.5} & 72.7 & \textbf{91.3} & 72.0 & 99.2 & \textbf{91.4} & \textbf{90.7} & \textbf{55.5} & \textbf{85.8} & \textbf{96.0} & \textbf{86.1} & \textbf{75.9} & \textbf{83.2} & \textbf{68.2} & \textbf{50.9} & \textbf{82.3} & \textbf{80.3} & \textbf{50.8} & \textbf{34.5} & \textbf{43.1}\\
    \bottomrule
  \end{tabular}}
  \caption{The detailed results of head-to-head comparison on VTAB-1k benchmark.}
  \label{table:vtab_appx}
\end{table*}

\begin{table*}[t]
  \centering
  \setlength{\tabcolsep}{2pt}
  \renewcommand{\arraystretch}{1.3}
  \begin{tabular}{ccc|cccccccc}
    \toprule
        Method & \#Params(M)& Avg. & MNLI & SST-2 & MRPC & CoLA & QNLI & QQP & RTE & STS-B \\
    \midrule
        LoRA           & 0.3   & 87.2 &  87.2 & 94.5  & 90.2  &  64.2 & 92.5 & 90.7 & 87.4 & 91.1 \\
        SeqLoRA       & 0.3   &  87.4& \textbf{87.5} & 94.7  & 90.7  &  63.3 & \textbf{93.0} & \textbf{90.8} & \textbf{88.1} & 91.4 \\
        Hydra               & 0.3   & \textbf{87.9}&  \textbf{87.5} & \textbf{95.0}  & \textbf{92.2}  &  \textbf{65.4} & 92.8 & \textbf{90.8} & 87.4 & \textbf{91.7} \\
    \bottomrule
  \end{tabular}
  \caption{The detailed results of head-to-head comparison on GLUE benchmark.}
  \label{table:nlu_ablation}
\end{table*}

\begin{table*}[ht]
    \centering
    \setlength{\tabcolsep}{2pt}
    \renewcommand{\arraystretch}{1.3}
    \resizebox{\textwidth}{!}{
    \begin{tabular}{ccc|cccccccccccccccccccc}
        \toprule
        Method & \rotatebox{90}{\#Params(M)} & \rotatebox{90}{Avg Acc.($\uparrow$)} & \rotatebox{90}{Caltech101} & \rotatebox{90}{CIFAR10} & \rotatebox{90}{CIFAR100} & \rotatebox{90}{Country211} & \rotatebox{90}{DTD} & \rotatebox{90}{EuroSAT} & \rotatebox{90}{FER2013} & \rotatebox{90}{FGVCAircraft} & \rotatebox{90}{Food101} & \rotatebox{90}{GTSRB} & \rotatebox{90}{HatefulMemes} & \rotatebox{90}{KITTI-Dist} & \rotatebox{90}{MNIST} & \rotatebox{90}{Flowers102} & \rotatebox{90}{OxfordPets} & \rotatebox{90}{PatchCamelyon} & \rotatebox{90}{SST2} & \rotatebox{90}{Resisc45} & \rotatebox{90}{StanfordCars} & \rotatebox{90}{VOC2007}\\ 
        \midrule
        
        MSA & 0.20 & 70.45 & \textbf{92.01} & 90.76 & 73.93 & 17.65 & \textbf{65.05} & 81.60 & \textbf{51.32} & 32.00 & \textbf{84.37} & 85.66 & 55.90 & \textbf{43.74} & \textbf{91.47} & 92.56 & 89.37 & 68.85 & \textbf{59.80} & 79.80 & 69.95 & \textbf{83.16} \\

        MLP & 0.20 & \textbf{70.95} & 91.23 & \textbf{90.89} & \textbf{74.20} & \textbf{17.75} & 64.47 & \textbf{87.00} & 51.10 & \textbf{33.05} & 84.27 & \textbf{87.11} & \textbf{55.91} & 42.05 & 90.76 & \textbf{93.18} & \textbf{89.38} & \textbf{70.83} & 59.58 & \textbf{82.41} & \textbf{71.19} & 82.66 \\
    
        \bottomrule
        
    \end{tabular}}
    \caption{The detailed results of the optimal position of \textit{Hydra} module experiment on ELEVATER benchmark.} 
    \label{table:position_hydra}
\end{table*}

\section{Experimental details}\label{appxB}
\subsection{Few-shot experiments}
Following ~\cite{he2022parameter}, we used SGD optimizer to fine-tune the model. We set $r_a = r_b = 2$ for the \textit{Hydra} module since the rank of the LoRA in this experiment is set to 4. We searched for the optimal learning rate and the weight decay using the automatic hyper-parameter tuning toolkit from ELEVATER~\cite{li2022elevater}. Every experiments of few-shot setting on the ELEVATER benchmark are conducted on a single NVIDIA-A100 GPU. We report the hyper-parameters in \cref{table:clip-few_hyper}.

\subsection{VTAB-1k experiments}
We used AdamW optimizer with cosine scheduler where warm-up epoch is set to 10. For the similar number of parameters compared with methods in benchmark, we set low ranks $r_a = r_b = 2$ in both MSA and MLP blocks. \textit{Hydra} is trained with 100 epochs for all datasets. Detailed hyper-parameters for each dataset are reported in \cref{table:vtab_hyper}. All of the results are conducted on a single NVIDIA-A100 GPU.  

\subsection{Natural language understanding experiments}\label{nlu_exp}
We used AdamW optimizer with a linear learning rate decay schedule and set warm-up iteration ratio as $0.06$. Across the datasets, the low ranks are set as $r_a = 4$ and $r_b = 8$. Following LoRA~\cite{hu2021lora}, the adaptation modules for the MRPC, RTE and STS-B experiments were initialized using the best MNLI experiment checkpoint. The hyper-parameters specific to each dataset are presented in Table~\ref{table:nlu_hyper}. We used 4 NVIDIA-A100 GPUs for training.

\begin{table*}[ht]
    \centering
    \setlength{\tabcolsep}{2pt}
    \renewcommand{\arraystretch}{1.3}
    \resizebox{\textwidth}{!}{
    \begin{tabular}{ccccccccccccccccccccc}
        \toprule
        Method & \rotatebox{90}{Caltech101} & \rotatebox{90}{CIFAR10} & \rotatebox{90}{CIFAR100} & \rotatebox{90}{Country211} & \rotatebox{90}{DTD} & \rotatebox{90}{EuroSAT} & \rotatebox{90}{FER2013} & \rotatebox{90}{FGVCAircraft} & \rotatebox{90}{Food101} & \rotatebox{90}{GTSRB} & \rotatebox{90}{HatefulMemes} & \rotatebox{90}{KITTI-Dist} & \rotatebox{90}{MNIST} & \rotatebox{90}{Flowers102} & \rotatebox{90}{OxfordPets} & \rotatebox{90}{PatchCamelyon} & \rotatebox{90}{SST2} & \rotatebox{90}{Resisc45} & \rotatebox{90}{StanfordCars} & \rotatebox{90}{VOC2007}\\ 
        \midrule
        
        Batch size      & 64    & 64    & 64    & 64    & 64    & 64    & 64    & 64    & 64    & 64    & 64    & 64    & 64    & 64    & 64    & 64    & 64    & 64    & 64    & 64 \\
        
        Epochs          & 50    & 50    & 50    & 50    & 50    & 50    & 50    & 50    & 50    & 50    & 50    & 50    & 50    & 50    & 50    & 50    & 50    & 50    & 50    & 50 \\
        
        Learning rate   & 1e-2  & 1e-3  & 1e-3  & 1e-3  & 1e-2   & 1e-2   & 1e-2   & 1e-2   & 1e-4 & 1e-1   & 1e-4    & 1e-2  & 1e-2   & 1e-2   & 1e-2    & 1e-2  & 1e-4    & 1e-2  & 1e-2  & 1e-1 \\
        
        Weight decay    & 3.16e-4    & 1.00e-6 & 1.00e+0 & 1.33e-6   & 1.00e-6  & 1.00e-6  & 1.00e-4  & 3.16e-6   & 1.00e-4  & 1.00e-2  & 1.00e+4  & 1.00e-6  & 1.00e-6  & 1.00e-6  & 1.00e-6  & 1.00e+0  & 1.00e-6  & 1.00e-4  & 3.16e-4   & 3.16e-1 \\
        \bottomrule
        
    \end{tabular}}
    \caption{The hyper-parameters in few-shot experiments. Since the weight decays for each dataset are searched in the logspace, we report them using two significant digits in the exponential notation.} 
    \label{table:clip-few_hyper}
\end{table*}

\begin{table*}[ht]
  \centering
  \setlength{\tabcolsep}{2pt}
  \begin{tabular}{cccccccccccccccccccc}
    \toprule
        Hyper-parameter & \rotatebox{90}{CIFAR100} & \rotatebox{90}{Caltech101} & \rotatebox{90}{DTD} & \rotatebox{90}{Flower102} & \rotatebox{90}{Pets} & \rotatebox{90}{SVHN} & \rotatebox{90}{Sun397} & \rotatebox{90}{Camelyon} & \rotatebox{90}{EuroSAT} & \rotatebox{90}{Resisc45} & \rotatebox{90}{Retinopathy} & \rotatebox{90}{Clevr-Count} & \rotatebox{90}{Clevr-Dist} & \rotatebox{90}{DMLab} & \rotatebox{90}{KITTI-Dist} & \rotatebox{90}{dSPR-Loc} & \rotatebox{90}{dSPR-Ori} & \rotatebox{90}{sNORB-Azim} & \rotatebox{90}{sNORB-Ele} \\
    \midrule
        Batch size      & 16 & 32 & 32 & 64 & 32 & 16 & 64 & 64 & 16 & 64 & 64 & 64 & 16 & 64 & 64 & 16 & 64 & 64 & 64 \\
        Learning rate   & 5e-4 & 1e-3 & 1e-3 & 5e-3 & 1e-3 & 1e-3 & 1e-3 & 5e-3 & 5e-4 & 5e-3 & 5e-4 & 5e-4 & 5e-4 & 1e-3 & 5e-4 & 1e-3 & 1e-3 & 5e-3 & 5e-4 \\
        Weight decay    & 1e-4 & 1e-4 & 1e-4 & 1e-4 & 1e-4 & 1e-4 & 1e-4 & 1e-4 & 1e-4 & 1e-4 & 1e-4 & 1e-4 & 1e-4 & 1e-4 & 1e-4 & 1e-4 & 1e-4 & 1e-5 & 1e-4 \\
        Dropout         & 0.2 & 0.1 & 0.2 & 0.0 & 0.0 & 0.0 & 0.2 & 0.2 & 0.0 & 0.2 & 0.2 & 0.2 & 0.0 & 0.2 & 0.2 & 0.2 & 0.2 & 0.2 & 0.2 \\
    \bottomrule
  \end{tabular}
  \caption{The hyper-parameters in VTAB-1k experiments.}
  \label{table:vtab_hyper}
\end{table*}

\begin{table*}[ht]
  \centering
  \setlength{\tabcolsep}{2pt}
  \renewcommand{\arraystretch}{1.3}
  \begin{tabular}{cccccccccc}
    \toprule
         Hyper-parameter & MNLI & SST-2 & MRPC & CoLA & QNLI & QQP & RTE & STS-B \\
    \midrule
         Batch size      & 8 & 16 & 16 & 4 & 8 & 16 & 16 & 4 \\
         Epochs          & 30 & 60 & 40 & 80 & 25 &25 & 80 & 40 \\
         Learning rate   & 4e-4 & 6e-4 & 6e-4 & 4e-4 & 8e-4 & 6e-4 & 8e-4 & 6e-4 \\
         Weight decay    & 0.1 & 0.1 & 0.1 & 0.1 & 0.1 & 0.1 & 0.2 & 0.1 \\
        
    \bottomrule
  \end{tabular}
  \caption{The hyper-parameters in NLU experiments.}
  \label{table:nlu_hyper}
\end{table*}

\section{Dataset details}\label{appxC}
\subsection{Few-shot experiments}\label{vision_data}

We tested Hydra on 20 datasets from ELEVATER benchmark~\cite{li2022elevater} for few-shot experiments: Caltech101, CIFAR10, CIFAR100, Country211, DTD, EuroSat, FER2013, FGVC Aircraft, Food101, GTSRB, HatefulMemes, KittiDistance, MNIST, Flowers102, OxfordPets, PatchCamelyon, SST2, RESISC45, StanfordCars, and VOC2007. Detailed statistics of each dataset are reported in \cref{table:vision_datasets}.

\subsection{VTAB-1k experiments}
\label{subsection:vtab_expdetail}
We assessed the accuracy using the VTAB-1k benchmark~\cite{zhai2019visual}, and the statistics of each dataset are presented in \cref{table:vtab_dataset}. The benchmark consists of various vision domain with three groups: CIFAR100, Caltech101, DTD, Flowers102, OxfordPets, SVHN and SUN397 in \textit{Natural} group, PatchCamelyon, EuroSAT, Resisc45 and Retinopathy in \textit{Specialized} group, and Clevr, DMLab, dSprites and SmallNORB in \textit{Structured} group. The number of train data is fixed to 1,000 for each dataset.

\subsection{Natural language understanding experiments}\label{nlu_data}
We evaluated our method on the General Language Understanding Evaluation (GLUE) benchmark~\cite{wang2018glue}. It covers various language understanding tasks, including linguistic acceptability (CoLA~\cite{warstadt2018neural}), sentiment analysis (SST-2~\cite{socher2013recursive}), paraphrase (MRPC~\cite{dolan2005automatically}, QQP\footnote{data.quora.com/First-Quora-Dataset-Release-Question-Pairs}), sentence similarity (STS-B~\cite{cer2017semeval}), and natural language inference (MNLI~\cite{williams2017broad}, QNLI~\cite{rajpurkar2016squad}, RTE~\cite{dagan2006pascal, haim2006second, giampiccolo2007third, bentivogli2009fifth}).

\begin{table*}[pt]
  \centering
  \setlength{\tabcolsep}{2pt}
  \renewcommand{\arraystretch}{1.2}
  \begin{tabular}{c @{\hspace{3em}} | @{\hspace{3em}} c @{\hspace{3em}} c @{\hspace{3em}}c} 
    \toprule
        Dataset                             & \#labels  & Train size & Test size \\ 
    \midrule
        Hateful Memes~\cite{kiela2020hateful}               & 2	        & 8,500	     & 500	\\
        PatchCamelyon~\cite{veeling2018rotation}	            & 2	        & 262,144    & 32,768 \\
        Rendered-SST2~\cite{radford2021learning}	            & 2	        & 6,920	     & 1,821	\\
        KITTI Distance~\cite{fritsch2013new}	            & 4	        & 6,347	     & 711	\\
        FER 2013~\cite{goodfellow2013challenges}	                & 7	        & 28,709	 & 3,589 \\
        CIFAR10~\cite{krizhevsky2009learning}	                    & 10	    & 50,000	 & 10,000 \\
        EuroSAT~\cite{helber2019eurosat}	                    & 10	    & 5,000	     & 5,000	\\
        MNIST~\cite{deng2012mnist}	                    & 10	    & 60,000	 & 10,000	\\
        STL10~\cite{coates2011analysis}	                    & 10	    & 5,000	     & 8,000 \\
        SVHN~\cite{goodfellow2013multi}	                    & 10	    & 73,257	 & 26,032 \\
        VOC 2007 Classification~\cite{everingham2010pascal}	    & 20	    & 2,501	     & 4,952 \\
        Oxford-IIIT-Pets~\cite{parkhi2012cats}	        & 37	    & 3,680	     & 3,669	\\
        GTSRB~\cite{stallkamp2011german}	                    & 43	    & 26,640	 & 12,630	\\
        Resisc45~\cite{cheng2017remote}	                & 45	    & 3,150	     & 25,200	\\
        Describable Textures~\cite{cimpoi2014describing}	    & 47	    & 1,880	     & 1,880 \\
        CIFAR100~\cite{krizhevsky2009learning}	                & 100	    & 50,000	 & 10,000 \\
        FGVC Aircraft~\cite{maji2013fine}	            & 100	    & 3,334	     & 3,333 \\
        Food101~\cite{bossard2014food}	                    & 101	    & 75,750	 & 25,250	\\
        Caltech101~\cite{fei2004learning}	                & 101	    & 3,060	     & 6,084 \\
        Oxford-Flowers102~\cite{nilsback2008automated}          & 102	    & 1,020	     & 6,149 \\
        Stanford Cars~\cite{krause20133d}	            & 196	    & 8,144	     & 8,041	\\
        Country-211~\cite{radford2021learning}	                & 211	    & 31,650	 & 21,100	\\
        SUN397~\cite{xiao2010sun}	                    & 397	    & 19,850	 & 19,850	\\
    \bottomrule
  \end{tabular}
  \caption{Statistics of 23 datasets used in few-shot experiments.}
  \label{table:vision_datasets}
\end{table*}

\begin{table*}[pt]
    \centering
    \setlength{\tabcolsep}{2pt}
    \renewcommand{\arraystretch}{1.2}
    \begin{tabular}{c@{\hspace{3em}}|@{\hspace{3em}}c@{\hspace{3em}}c@{\hspace{3em}}c}
        \toprule
            Dataset & \#labels & Train size & Test size \\
        \midrule
            CIFAR100~\cite{krizhevsky2009learning} & 100 &  & 10,000 \\
            Caltech101~\cite{fei2004learning} & 102 &  & 6,084 \\
            Describable Textures~\cite{cimpoi2014describing} & 47 & & 1,880 \\
            Oxford-Flowers102~\cite{nilsback2008automated} & 102 & & 6,149 \\
            Oxford-IIIT-Pets~\cite{parkhi2012cats} & 37 & & 3,669 \\
            SVHN~\cite{goodfellow2013multi} & 10 & & 26,032 \\
            Sun397~\cite{xiao2010sun} & 397 & & 36,032 \\
            PatchCamelyon~\cite{veeling2018rotation} & 2 & & 32,768\\
            EuroSAT~\cite{helber2019eurosat} & 10 & & 5,400\\
            Resisc45~\cite{cheng2017remote} & 45 & 1,000 & 6,300 \\
            Retinopathy~\cite{graham2015kaggle} & 5 & & 42,670 \\
            Clevr/count~\cite{johnson2017clevr} & 8 & & 15,000\\
            Clevr/distance~\cite{johnson2017clevr} & 6 & & 15,000\\
            DMLab~\cite{beattie2016deepmind} & 6 & & 22,735\\
            KITTI Distance~\cite{fritsch2013new} & 4 & & 711\\
            dSprites/location~\cite{matthey2017dsprites} & 16 & & 73,728 \\
            dSprites/orientataion~\cite{matthey2017dsprites} & 16 & & 73,728\\
            SmallNORB/azimuth~\cite{lecun2004learning} & 18 & & 12,150 \\
            SmallNORB/elevation~\cite{lecun2004learning} & 18 & & 12,150 \\
        \bottomrule
    \end{tabular}
    \caption{Datasets of VTAB-1k benchmark}
    \label{table:vtab_dataset}
\end{table*}


\end{document}